\documentclass[10pt,twocolumn,letterpaper]{article}

\usepackage{subcaption}

\usepackage{iccv}
\usepackage{times}
\usepackage{epsfig}
\usepackage{graphicx}
\usepackage{amsmath}
\usepackage{amssymb}
\usepackage{enumerate}
\usepackage{makecell}

\usepackage{multirow}

\usepackage[dvipsnames]{xcolor}

\newlength\savewidth
\newcommand\Tstrut{\rule{0pt}{2.5ex}}   
\newcommand\Bstrut{\rule[-1ex]{0pt}{0pt}}
\newcommand\TBstrut{\Tstrut\Bstrut}

\usepackage[pagebackref=true,breaklinks=true,letterpaper=true,colorlinks,bookmarks=false]{hyperref}

\arxivcopy

\begin{document}

%%%%%%%%% TITLE
\title{AutoLoss-Zero: Searching Loss Functions from Scratch for Generic Tasks}

\author{Hao Li$^{1*\dag}$, Tianwen Fu$^{2}$\thanks{\noindent Equal contribution. $^{\dag}$This work is done when Hao Li and Tianwen Fu are interns at SenseTime Research. $^{\ddag}$Corresponding author.}~~$^{\dag}$, Jifeng Dai$^{3,5}$, Hongsheng Li$^{1}$, Gao Huang$^{4}$, Xizhou Zhu$^{3\ddag}$
\vspace{0.3em}\\
$^{1}$CUHK-SenseTime Joint Laboratory, The Chinese University of Hong Kong \\ 
$^{2}$Department of Information Engineering, The Chinese University of Hong Kong \\
$^{3}$SenseTime Research \quad $^{4}$Tsinghua University \\
$^{5}$Qing Yuan Research Institute, Shanghai Jiao Tong University \\
\texttt{\small haoli@link.cuhk.edu.hk},~ \texttt{\small futianwen@ie.cuhk.edu.hk} \\
\texttt{\small \{daijifeng, zhuwalter\}@sensetime.com}
\\
\texttt{\small hsli@ee.cuhk.edu.hk},~ \texttt{\small gaohuang@tsinghua.edu.cn}
% \vspace{-1em}
}

\maketitle
\thispagestyle{empty}

\def\Name{AutoLoss-Zero}

%%%%%%%%% ABSTRACT
\begin{abstract}

Significant progress has been achieved in automating the design of various components in deep networks. However, the automatic design of loss functions for generic tasks with various evaluation metrics remains under-investigated. Previous works on handcrafting loss functions heavily rely on human expertise, which limits their extendibility. Meanwhile, existing efforts on searching loss functions mainly focus on specific tasks and particular metrics, with task-specific heuristics. Whether such works can be extended to generic tasks is not verified and questionable. In this paper, we propose \Name, the first general framework for searching loss functions from scratch for generic tasks. Specifically, we design an elementary search space composed only of primitive mathematical operators to accommodate the heterogeneous tasks and evaluation metrics. A variant of the evolutionary algorithm is employed to discover loss functions in the elementary search space. A loss-rejection protocol and a gradient-equivalence-check strategy are developed so as to improve the search efficiency, which are applicable to generic tasks. Extensive experiments on various computer vision tasks demonstrate that our searched loss functions are on par with or superior to existing loss functions, which generalize well to different datasets and networks. Code shall be released.

\end{abstract}

%%%%%%%%% BODY TEXT
\section{Introduction}

Recent years have witnessed exciting progress in AutoML for deep learning~\cite{zoph2016neural,pham2018efficient,liu2018darts,liu2020evolving,pham2021autodropout,cubuk2019autoaugment}.
The automatic design of many components has been explored, ranging from architectures (\eg, neural architectures~\cite{ren2020comprehensive} and normalization-activation operations~\cite{liu2020evolving}) to learning strategies (\eg, data augmentation strategies~\cite{cubuk2019autoaugment}, dropout patterns~\cite{pham2021autodropout}, and training hyper-parameters~\cite{dai2020fbnetv3}).
However, to automate the entire deep learning process, an essential component is under-investigated, namely, the automatic design of loss functions for generic tasks.

Loss functions are indispensable parts in deep network training. In various tasks, including semantic segmentation~\cite{chen2017deeplab,zhao2017pyramid}, object detection~\cite{girshick2015fast,ren2015faster}, instance segmentation~\cite{he2017mask,cai2019cascade} and pose estimation~\cite{sun2019deep}, cross-entropy (CE) and L1/L2 losses are the default choices for categorization and regression, respectively. As the default loss functions are usually approximations for specific evaluation metrics, there usually exists a misalignment between the surrogate loss and the final evaluation metric. For example, for bounding box localization in object detection, L1 loss is widely used, while the IoU metric is the standard evaluation metric~\cite{yu2016unitbox}. Similar discrepancy has also been observed in semantic segmentation~\cite{li2020auto}, where some metrics measure the accuracy of the whole image, while others focus more on the segmentation boundaries. The misalignment between network training and evaluation results in sub-optimal solutions with degraded performance.

A multitude of handcrafted loss functions have been proposed for different evaluation metrics. Since most desired metrics are non-differentiable and cannot be directly used as training objectives, many existing works~\cite{ronneberger2015u,wu2016bridging,lin2017focal,li2019gradient,caliva2019distance,qin2019basnet,girshick2015fast} design differentiable variants of the CE and L1/L2 losses by carefully analyzing specific evaluation metrics. Another series of works~\cite{berman2018lovasz,milletari2016v,rahman2016optimizing,yu2016unitbox,rezatofighi2019generalized,zheng2020distance,qian2020dr} handcraft clever surrogate losses based on the mathematical expressions of specific evaluation metrics. Although these handcrafted loss functions show improvement on their target metrics, they heavily rely on expertise and careful analysis for specific scenarios, which limits their extendibility.

In this paper, we aim to automate the process of designing loss functions for generic tasks. Although there are several pioneer works~\cite{li2019lfs,wang2020loss,liu2021loss,li2020auto} on loss function search, they are all limited to specific tasks and particular evaluation metrics, with task-specific heuristics. Whether such task-specific heuristics can be applied to generic tasks is not verified and questionable. Searching loss functions for generic tasks is much more challenging, because of the heterogeneity of various tasks and evaluation metrics. 
The search space should be composed of basic primitive operators so as to accommodate such heterogeneity, and the search algorithm should be efficient enough so as to find the best combination of the basic primitives for the given task and evaluation metric. Meanwhile, no task-specific heuristics should be involved in the search algorithm.

This paper presents the first general loss function search framework applicable to various evaluation metrics across different tasks, named \Name. We build our search space only with primitive mathematical operators to enjoy the high diversity and expressiveness. A variant of the evolutionary algorithm is employed to discover the high-quality loss functions from scratch with minimal human expertise. Specifically, \Name\ formulates loss functions as computational graphs composed only of primitive mathematical operators (see Table~\ref{Methods:Operators}). The computation graphs are randomly built from scratch, and are evolved according to their performance on the target evaluation metrics. In the search algorithm,  to improve the search efficiency, we propose a loss-rejection protocol that efficiently filters out the unpromising loss function candidates, which brings great speed-up to the search procedure. A gradient-equivalence-check strategy is developed to avoid duplicate evaluations of equivalent loss functions. The loss-rejection protocol and the gradient-equivalence-check strategy are generally applicable to various tasks and metrics.

We validate our framework on various computer vision tasks, including semantic segmentation, object detection, instance segmentation, and pose estimation. Extensive experiments on large-scale datasets such as COCO~\cite{lin2014microsoft}, Pascal VOC~\cite{everingham2015pascal} and Cityscapes~\cite{cordts2016cityscapes} show that the searched losses are on par with or superior to existing handcrafted and specifically searched loss functions. Ablation studies show that our searched loss functions can effectively generalize to different networks and datasets.

Our main contributions can be summarized as follows:
\begin{itemize}
    \vspace{-0.5em}
    \item \Name\ is the first general AutoML framework to search loss functions from scratch for generic tasks with minimal human expertise. The effectiveness is demonstrated on a variety of computer vision tasks.
    \vspace{-0.5em}
    \item A novel loss-rejection protocol is developed to filter out the unpromising loss functions efficiently. A gradient-equivalence-check strategy is also developed to avoid duplicate evaluations. These techniques bring great improvement to the search efficiency, and are generally applicable to various tasks and metrics.
    \vspace{-0.5em}   
    \item The searched loss functions by themselves are contributions, because they are transferable across different models and datasets with competitive performance. 
\end{itemize}

\section{Related Work}

\noindent\textbf{Hand-crafted loss functions} for prevalent evaluation metrics have been studied by numerous works.
A large fraction of previous works develop loss function variants based on the standard cross-entropy loss and L1/L2 loss. For categorization, \cite{ronneberger2015u,wu2016bridging,lin2017focal,li2019gradient} mitigate the imbalance of samples by incorporating different sample weights. \cite{caliva2019distance,qin2019basnet} propose to up-weight the losses at boundary pixels to deliver more accurate boundaries. For regression, Smooth-L1 loss~\cite{girshick2015fast} is proposed for improved stability and convergence. 
Another line of research~\cite{berman2018lovasz,milletari2016v,rahman2016optimizing,yu2016unitbox,rezatofighi2019generalized,zheng2020distance,zheng2020enhancing,qian2020dr} deals with the misalignment between loss functions and various evaluation metrics by handcrafting differentiable extensions or surrogates of metrics as loss functions, including segmentation IoU~\cite{berman2018lovasz,rahman2016optimizing}, F1 score~\cite{milletari2016v}, bounding box IoU~\cite{yu2016unitbox,rezatofighi2019generalized,zheng2020distance,zheng2020enhancing}, and Average Precision~\cite{qian2020dr}.

Although these handcrafted losses are successful under different scenarios, they heavily rely on careful design and expertise for analyzing the property of specific metrics. In contrast, we propose an automated loss design framework that is generally suitable for different tasks and metrics.

\vspace{0.5em}
\noindent\textbf{Direct optimization} for non-differentiable evaluation metrics has also been studied.
For structural SVMs~\cite{tsochantaridis2005large}, \cite{joachims2005support,yue2007support,ranjbar2012optimizing} propose non-gradient methods to directly optimize ideal metrics. \cite{hazan2010direct,song2016training,mohapatra2018efficient} apply loss-augmented inference to derive the gradients from the expectation of metrics. However, the computational complexity is high, which requires specifically designed efficient algorithms for different metrics. 
Policy gradients~\cite{rao2018learning,ranzato2015sequence,wu2017sequence,bahdanau2016actor,shen2015minimum} are also adopted to directly optimize non-differentiable metrics. However, these methods suffer from: 1) complicated action space, which requires task specific approximations~\cite{rao2018learning}; 2) high variance of gradient estimation and objective instability~\cite{wu2018study}.

Recently, \cite{chen2020ap,oksuz2020ranking} adopt error-driven learning for object detection, which is limited to specific scenarios.

Although these direct optimization methods mitigate the mis-alignment issue between training objectives and evaluation metrics, they require specific analysis and designs for the target metrics.

\vspace{0.5em}
\noindent\textbf{AutoML for generic tasks} has long been pursued in machine learning research~\cite{he2019automl}. 
Recent works include automated search for neural architecture (NAS)~\cite{zoph2016neural,pham2018efficient,liu2018darts}, normalization-activation operations~\cite{liu2020evolving}, dropout patterns~\cite{pham2021autodropout}, data augmentation~\cite{cubuk2019autoaugment}, and training hyper-parameters~\cite{dai2020fbnetv3}. 
Most of the existing works aim to specialize an architecture built upon expert-designed operators~\cite{zoph2016neural,pham2018efficient,liu2018darts}, or search for specific hyper-parameters in a fixed formula~\cite{pham2021autodropout,cubuk2019autoaugment,dai2020fbnetv3}.

Our work shares a similar philosophy to AutoML-Zero~\cite{real2020automl} and EvoNorm~\cite{liu2020evolving}, which employ evolutionary algorithms to search for ML algorithms or normalization-activation operations from only primitive mathematical operations. 
However, for loss functions, the search space design  is quite different and there are unique properties that can be leveraged for efficient search. We introduce 1) an effective search space for loss functions with specific initialization and mutation operations; and 2) a loss-rejection protocol and a gradient-equivalence-check strategy to improve the search algorithm efficiency.

\vspace{0.5em}
\noindent\textbf{Loss function search} has also raised the interest of researchers in recent years. All the pioneer works~\cite{li2019lfs,wang2020loss,liu2021loss,li2020auto} are limited to specific tasks and specific metrics, with task-specific heuristics. Specifically, \cite{li2019lfs,wang2020loss} search for optimal losses for face recognition. The searched loss functions are optimal combinations of existing handcrafted variants of the cross-entropy loss. As the resulting objective is essentially an integration of existing loss functions, it cannot solve the mis-alignment between cross-entropy losses and many target metrics well. Recently, \cite{li2020auto} proposes to search loss functions for semantic segmentation by substituting the logical operations in metrics with parameterized functions. However, such parameterization cannot be easily extended for generic metrics, such as mAP in object detection, where the matching and ranking processes are difficult to be parameterized. Another closely related work is \cite{liu2021loss}, which searches loss functions for object detection. Similar to our method, \cite{liu2021loss} also formulates loss functions as the combination of primitive operators. However, \cite{liu2021loss} initializes the search from well-performed handcrafted loss functions specific for object detection, and separately searches for one loss branch with the other loss branch fixed as initialization. Moreover, \cite{liu2021loss} designs their loss-rejection protocol specifically for object detection, and cannot be applied to other tasks. In contrast, our method can simultaneously search for multiple loss branches from random initialization without starting from any human-designed loss functions. Our method has no specialized design for specific tasks or metrics, and consequently is applicable to generic tasks.

\section{Method}

Given a task (\eg, semantic segmentation and object detection) and a corresponding evaluation metric (\eg, mIoU and mAP), \Name\ aims to automatically search a proper loss function from scratch for training a neural network. 
A general search space is proposed, in which each loss function is represented as a computational graph.
The graph takes the network predictions and ground truths as inputs, and transforms them into a final loss value.
With minimal human expertise, only primitive mathematical operations (see Table~\ref{Methods:Operators}) are used as the intermediate computational nodes to accommodate the high diversity among different tasks and metrics.
An efficient evolutionary algorithm is employed to search the loss function for the given task and metric. 
To enable the evolution, effective random initialization and mutation operations are defined. A novel loss-rejection protocol and gradient-equivalence-check strategy are also proposed to improve the search efficiency, which are applicable to generic tasks.

\setlength{\tabcolsep}{4pt}
\renewcommand{\arraystretch}{1.2}
\begin{table}[t]
\begin{center}
\small
\resizebox{0.85\linewidth}{!}{
\begin{tabular}{llc}
\hline
\textbf{Element-wise Operator} & \textbf{Expression} & \textbf{Arity}\TBstrut \\
\hline
Add & $x + y$ & 2\Tstrut\\
Mul & $x \times y$ & 2\Bstrut \\
\hline
Neg & $-x$ & 1\Tstrut \\
Abs & $|x|$ & 1 \\
Inv & $1 / (x + \epsilon)$ & 1 \\
Log & $\text{sign}(x) \cdot \log(|x| + \epsilon)$ & 1 \\
Exp & $e^x$ & 1 \\
Tanh & $\tanh(x)$ & 1 \\
Square & $x^2$ & 1 \\
Sqrt & $\text{sign}(x) \cdot \sqrt{|x| + \epsilon}$ & 1\Bstrut\\
\hline
\textbf{$^\dag$Aggregation Operator} & \textbf{Expression} & \textbf{Arity}\TBstrut \\
\hline
Mean$_{nhw}$ & $\frac{1}{NHW} \sum_{nhw} x_{nchw}$ & 1\Tstrut\\
Mean$_{c}$ & $\frac{1}{C} \sum_{c} x_{nchw}$ & 1\\
Max-Pooling$_{3\times3}$ & $\mathrm{Max\text{-}Pooling}_{3\times3}(x)$ & 1 \\
Min-Pooling$_{3\times3}$ & $\mathrm{Min\text{-}Pooling}_{3\times3}(x)$ & 1\Bstrut\\
\hline
\end{tabular}}
\end{center}
\caption{Primitive operator set $\mathcal{H}$. $x$ and $y$ are of the same shape of $(N,C,H,W)$, which are the input tensors of the operators. $\epsilon = 10^{-12}$ is a small positive number for avoiding infinite values or gradients. This primitive operator set is shared in all of our experiments. $\dag$ Each aggregation operator is a mapping that replaces the elements of the input tensor with the aggregated values. Both the stride and padding of Max/Min-Pooling are set as $1$. Thus, all of the operators preserve the shape of the input tensor(s).}
\vspace{-0.5em}
\label{Methods:Operators}
\end{table}

\begin{figure*}[t]
    \begin{center}
        \includegraphics[width=0.9\textwidth]{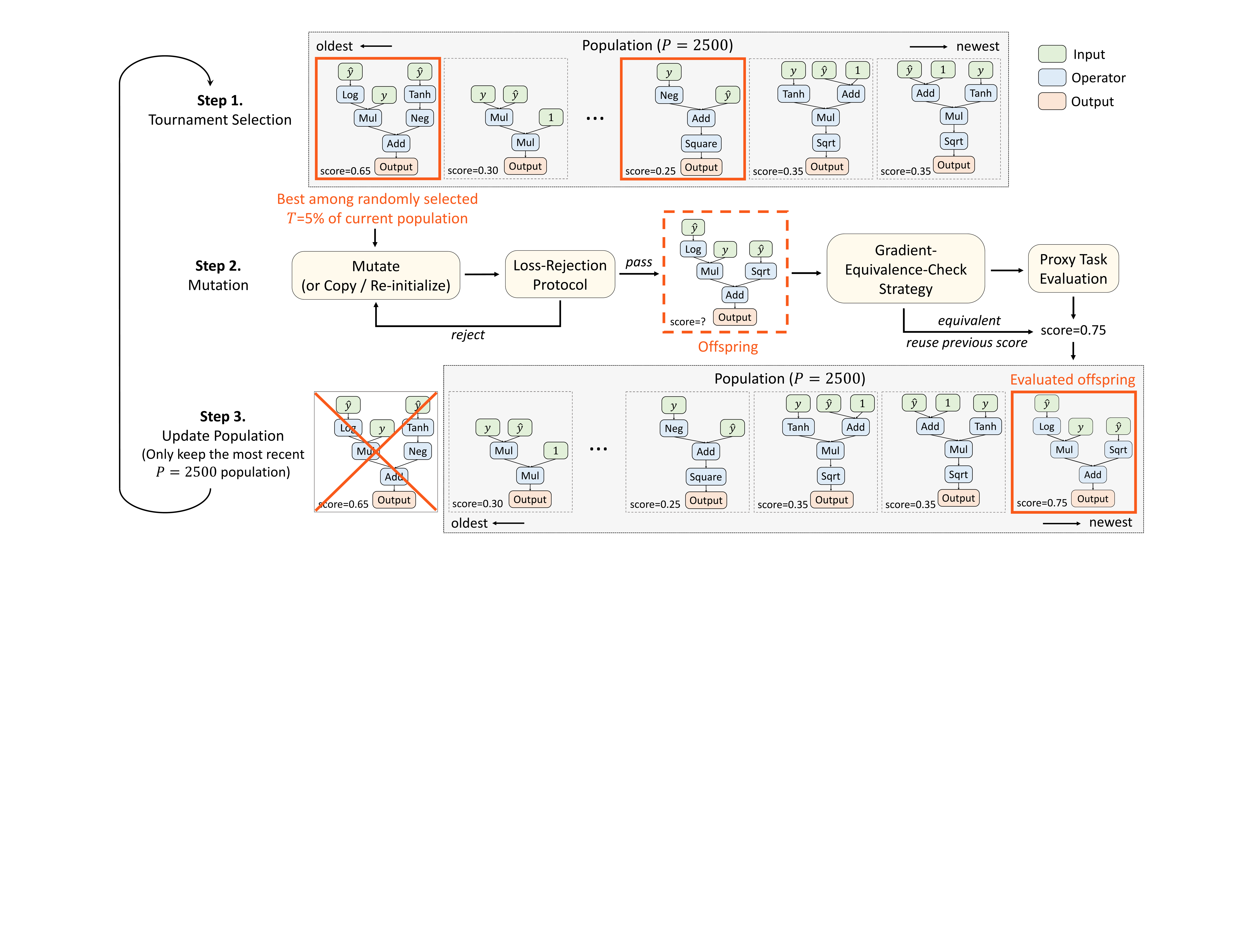}
    \end{center}
    \vspace{-0.5em}
    \caption{Overview of the search pipeline.}
    \vspace{-1.0em}
    \label{fig:Methods:Pipeline}
\end{figure*}

\subsection{Search Space}

The search spaces of most AutoML approaches~\cite{zoph2016neural,pham2018efficient,liu2018darts,liu2020evolving,pham2021autodropout,cubuk2019autoaugment} are specially designed for particular purposes and not suitable for loss functions. In loss function search, \cite{li2020auto} proposes a loss function search space specifically for semantic segmentation, which cannot be extended to generic tasks. In \cite{li2019lfs,wang2020loss}, the search space is simply the combination of existing loss functions, which cannot form new loss functions. The search space of \cite{liu2021loss} is also of primitives, which is most similar to ours. However, the primitives of \cite{liu2021loss} is a constrained set for the specific task of object detection. In this subsection, we design a general search space for loss functions applicable for generic tasks and evaluation metrics.

\Name\ seeks to search the proper loss function for training the networks that maximizes the given evaluation metric $\xi$. The loss function $L(\hat{y}, y; \mathcal{N}_\omega)$ is defined on the network prediction $\hat{y}$ and its ground-truth training target $y$, where $\mathcal{N}_\omega$ is a network parameterized with $\omega$. The search target can be formulated as a nested optimization,
\begin{align}
\max_L \quad & f(L;\xi) = \xi\left(\mathcal{N}_{\omega^*(L)}; \mathcal{S_\text{eval}}\right), \\
\text{s.t. } \quad & \omega^*(L) = \arg \min_\omega~ \mathop{\mathbb{E}}\nolimits_{(\hat{y}, y) \in \mathcal{S_\text{train}}}\left[ L(\hat{y}, y; \mathcal{N}_\omega) \right],
\end{align}
where $f(L;\xi)$ is the evaluation score of the loss function $L$ under the given metric $\xi$, and $\omega^*(L)$ is the network parameters trained with $L$. $\mathbb{E}[\cdot]$ is the mathematical expectation. $\mathcal{S_\text{train}}$ and $\mathcal{S_\text{eval}}$ are the training and evaluation sets used in the search process, respectively.
The network prediction $\hat{y}$ and its training target $y$ share the same shape of $(N, C, H, W)$. For each tensor, we use $N, C, H, W$ to refer to the size of its batch, channel, width and height, respectively\footnote{For the predictions and training targets without spatial dimensions, we set $H=1$ and $W=1$ without loss of generality.}.

\vspace{0.5em}
\noindent\textbf{Loss Function Representation.~} 
The loss function $L$ is represented as a computational graph $\mathcal{G}$. The computational graph is a rooted tree, where the leaf nodes are inputs (\ie, network predictions and training targets), and the root is the output. The intermediate computational nodes are selected from a set of primitive mathematical operations (see Table~\ref{Methods:Operators}), which transform the inputs into the final loss value. 

The input tensors of the computational graph are sampled with replacement from $\{y, \hat{y}, 1\}$, where the additional constant $1$ is included to improve the flexibility of the search space.
The output tensor $o$ has the same shape of $(N, C, H, W)$ as the inputs, which is further aggregated to form the final loss value as
\begin{equation}
L(\hat{y}, y) = \frac{1}{NHW} \sum\nolimits_{nchw} o_{nchw}.
\end{equation}
Here, we do not normalize among the channel dimension, following the common practice of the cross-entropy loss.

As some tasks may have multiple loss branches (\eg, the classification and regression branches in object detection), we represent the loss of each branch as an individual computational graph, and sum their loss values together as the final loss. 
For a loss with $M$ branches, given the predictions $\{\hat{y}^1, \hat{y}^2, \dots,\hat{y}^M\}$ and their ground-truth training targets $\{y^1, y^2, \dots,y^M\}$ of each loss branch, the final loss function is represented as
\begin{equation}
L(\hat{y}, y) = \sum\nolimits_{i=1}^{M} L_i(\hat{y}^i, y^i).
\end{equation}

\vspace{0.5em}
\noindent\textbf{Primitive Operators.~} 
Table~\ref{Methods:Operators} summarizes the primitive operator set $\mathcal{H}$ used in our search space, including element-wise operators and aggregation operators that enable information exchange across spatial and channel dimensions.
Each aggregation operator is a mapping that replaces the elements of the input tensor with the aggregated values.
All the primitive operators preserve the shape of the input tensors in order to ensure the validity of computations.

\subsection{Search Algorithm}
\label{Sec:Methods.Search}

Inspired by the recent applications of AutoML~\cite{real2020automl,liu2020evolving}, a variant of evolutionary algorithm is employed for searching the proper loss functions. In existing works of loss function search \cite{li2019lfs,wang2020loss,li2020auto,liu2021loss}, variants of reinforcement learning or evolution algorithm are also adopted. However, the search methods in \cite{li2019lfs,wang2020loss,li2020auto,liu2021loss} are designed for searching in specific tasks and particular metrics, with task-specific heuristics. Whether such heuristics can be applied to generic tasks is not verified and questionable. Here, \Name\ searches loss functions for generic tasks from random initialization with minimal human expertise. The proposed method has no specialized design for specific tasks or metrics, which is widely applicable to generic tasks.

Figure~\ref{fig:Methods:Pipeline} illustrates the search pipeline of \Name. At initialization, $K$ loss functions ($K=20$ by default) are randomly generated to form the initial population. Each evolution picks $T$ ratio of population ($T = 5\%$ by default) at random, and selects the one with the highest evaluation score as the parent, \ie, tournament selection~\cite{goldberg1991comparative}. The parent is used to produce offspring through well-designed mutation operations. Following \cite{liu2020evolving,real2020automl}, only the most recent $P$ loss functions ($P=2500$ by default) are maintained.

As the search space is very sparse with a large number of unpromising loss functions, a novel loss-rejection protocol is developed to efficiently filter out loss functions that are not negatively correlated with the given evaluation metric. During the search, the initialization~/~mutation process of an individual loss function would be repeated until the resulting loss function can pass the loss-rejection protocol.

In order to further improve the search efficiency, a gradient-equivalence-check strategy is developed to avoid re-evaluating mathematically equivalent loss functions. 
Similar to other AutoML works~\cite{liu2020evolving,real2020automl}, lightweight proxy tasks are employed to 
reduce the computational cost of evaluating loss functions, which will be discussed at the end of this subsection.

\begin{figure}[t]
    \begin{center}
        \includegraphics[width=0.9\linewidth]{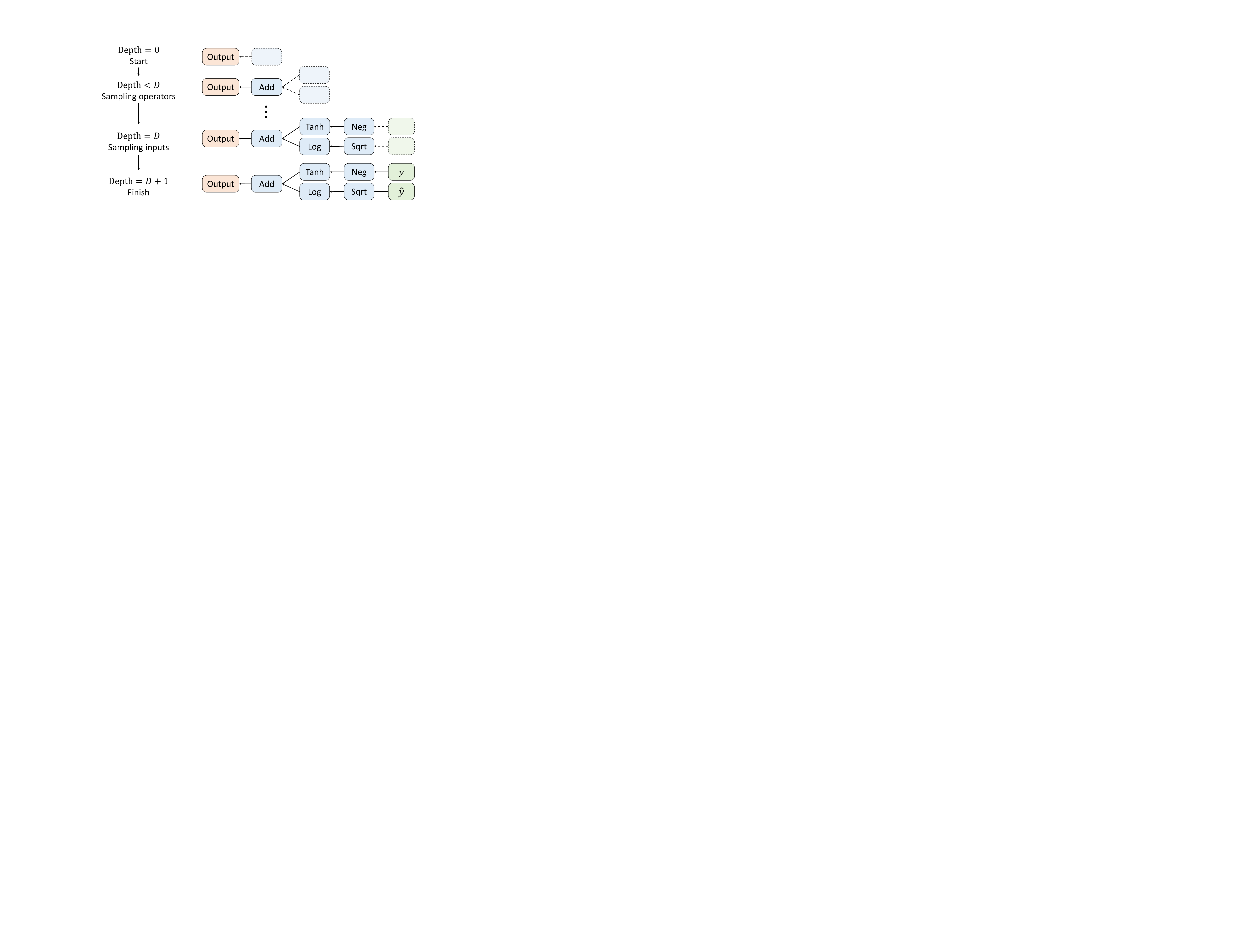}
    \end{center}
    \vspace{-0.5em}
    \caption{Random initialization of loss functions.}
    \label{fig:Methods:Generation}
\end{figure}

\vspace{0.5em}
\noindent\textbf{Random Initialization of Loss Functions.~} 
The computational graph of each initial loss function is randomly generated. Figure~\ref{fig:Methods:Generation} illustrates the process of loss function generation.
Starting from a graph with root (\ie, the output node) only, each node would randomly sample one or two operators from the primitive operator set $\mathcal{H}$ (see Table~\ref{Methods:Operators}), and append to the graph as its child node(s).
The root has one child. For each computational node, the number of child nodes is decided by its operator arity. 

When a computational node reaches the target depth $D$ ($D = 3$ by default), it randomly selects input tensor(s) as its child node(s). The input tensors would be the leaf nodes of the computational graph. Each randomly generated computational graph has a depth of $D + 1$, with $D$ computational nodes on each path from the root to a leaf node.

\begin{figure}[t]
    \begin{center}
        \includegraphics[width=1.0\linewidth]{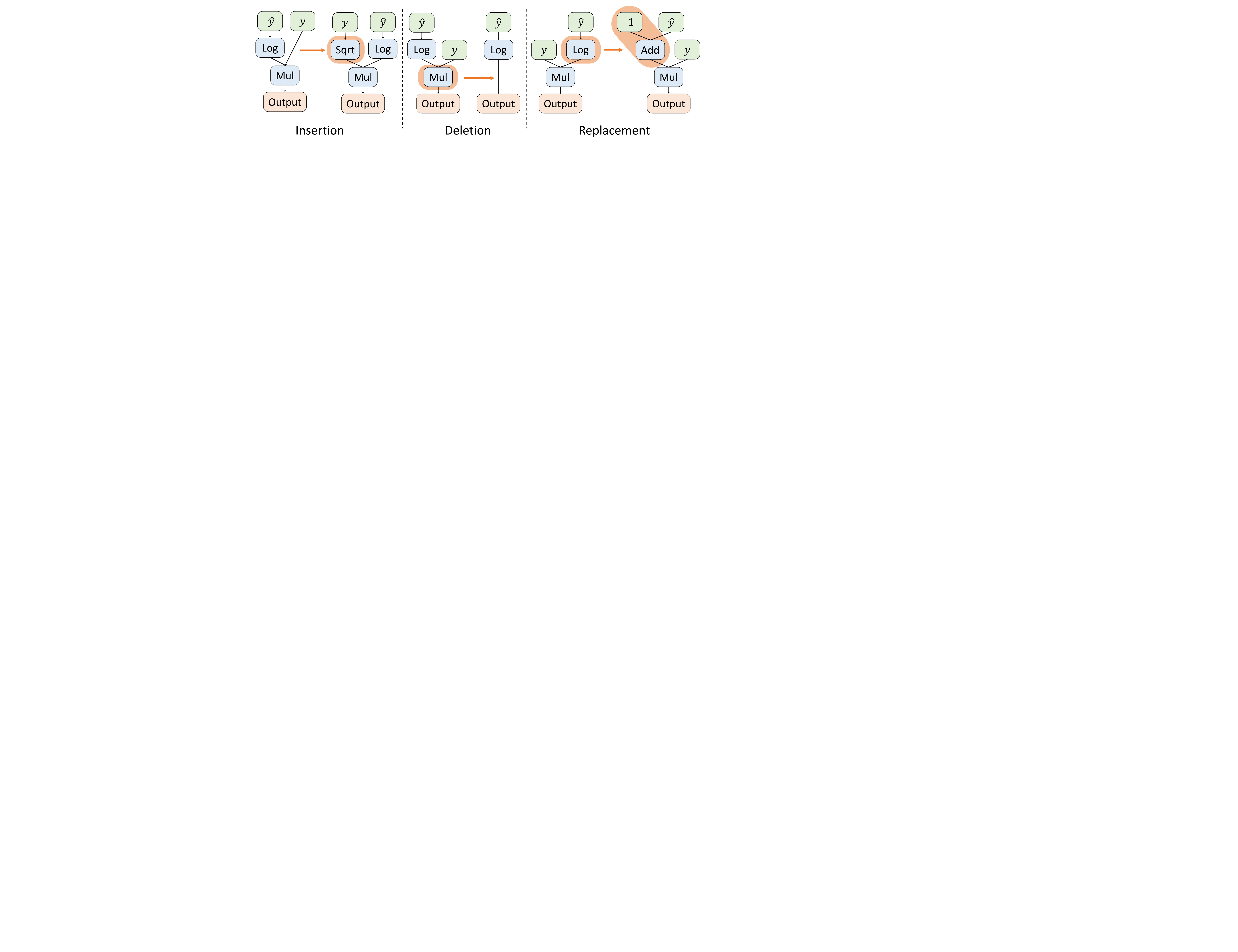}
    \end{center}
    \vspace{-0.5em}
    \caption{Candidate mutation operations.}
    \vspace{-1.0em}
    \label{fig:Methods:Mutation}
\end{figure}

\vspace{0.5em}
\noindent\textbf{Mutation.~}
The mutation process is inspired by \cite{real2020automl}, but the candidate mutation operations are specially designed for our search space. Figure~\ref{fig:Methods:Mutation} illustrates the candidate mutation operations, which are defined as:
\begin{itemize}
\vspace{-0.5em}
\item \emph{Insertion.} An operator randomly sampled from $\mathcal{H}$ is inserted between a randomly selected non-root node and its parent. If the operator has an arity of 2, it would randomly select an input as the additional child.
\vspace{-0.5em}
\item \emph{Deletion.} An intermediate computational node is randomly selected and removed. For the removed node, one of its child nodes is randomly picked to become the new child of its parent.
\vspace{-0.5em}
\item \emph{Replacement.} An operator is randomly sampled from $\mathcal{H}$ to replace a randomly selected non-root node. If the non-root node has children more than the arity of the operator, a random subset of the child nodes with the same number as the arity are kept as the children of the operator. Otherwise, it would randomly select inputs as the additional child nodes.
\vspace{-0.5em}
\end{itemize}
To produce the offspring, the given computational graph is processed by the three sequential steps:
\begin{enumerate}
\vspace{-0.5em}
\item Directly copy the graph with a probability of 10\%.
\vspace{-0.5em}
\item If copying is not performed, randomly re-initialize the full computational graph with a probability of 50\%.
\vspace{-0.5em}
\item If re-initialization is not performed, sequentially perform two mutation operations uniformly sampled from \{\emph{Insertion}, \emph{Deletion}, \emph{Replacement}\}.
\vspace{-0.5em}
\end{enumerate}

\vspace{0.5em}
\noindent\textbf{Loss-Rejection Protocol.~}
Our search space is highly flexible, in which only primitive operations are used to construct the loss functions. Similar to \cite{liu2020evolving,real2020automl,liu2021loss}, such flexibility leads to a large and sparse search space. Most loss function candidates result in network performance that is not better than random guessing. 
In loss function search, to improve the search efficiency, \cite{liu2021loss} designs a loss-rejection protocol to filter out unpromising loss functions before training the networks. However, it is specifically designed for object detection, which cannot be directly applied to generic tasks. 
Here, we propose a novel loss-rejection protocol that is generally applicable to various tasks and metrics.

Inspired by the fact that minimizing the proper loss functions should correspond to maximizing the given evaluation metric, we develop an efficient loss-rejection protocol for generic tasks.
Given $B$ random samples ($B = 5$ by default) from the training set $\mathcal{S_\text{train}}$ and a randomly initialized network $\mathcal{N}_{\omega_0}$, we record the network predictions and the corresponding training targets as $\{(\hat{y}_{b}, y_{b})\}_{b=1}^{B}$.
To efficiently estimate the correlation between the given evaluation metric $\xi$ and a candidate loss function $L$, a correlation score $g(L; \xi)$ is calculated as
\begin{align}
& g(L; \xi) = \frac{1}{B} \sum\nolimits_{b=1}^{B} \xi\left(\hat{y}_b^*(L), y_b\right) - \xi\left(\hat{y}_b, y_b\right), \\
& \text{s.t. } \quad \hat{y}_b^*(L) = \arg \min\nolimits_{\hat{y}_b}~ L(\hat{y}_b, y_b), \label{Eq:Method.EarlyReject}
\end{align}
where $\hat{y}_b^*(L)$ is the network predictions optimized with loss function $L$.
A large $g(L; \xi)$ indicates that minimizing the loss function $L$ corresponds to maximizing the given evaluation metric $\xi$. Otherwise, if $g(L; \xi)$ is less than a threshold $\eta$, the loss function $L$ is regarded as unpromising, which should be rejected.

Here, to speed up the rejection process, the loss function optimization is directly applied to the network prediction $\hat{y}_b$, instead of the network parameters $\omega$. Since the network computation is omitted, the rejection process is very efficient. With a single GPU, the proposed loss-rejection protocol can reach a throughput of 500$\sim$1000 loss functions per minute. In search, the initialization~/~mutation process of an individual loss function would be repeated until the resulting loss function can pass the loss-rejection protocol.

\vspace{0.5em}
\noindent\textbf{Gradient-Equivalence-Check Strategy.~}
To avoid re-evaluating mathematically equivalent loss functions, a gradient-equivalence-check strategy is developed. 
For each loss function $L$, we compute its gradient norms w.r.t. the network predictions used in the loss-rejection protocol as $\left\{{\left\|\partial L / \partial \hat{y}_b\right\|}_2\right\}_{b=1}^B$. 
If for all of the $B$ samples, two loss functions have the same gradient norms within two significant digits, they are considered equivalent, and the previous evaluation metric score would be reused.

\vspace{0.5em}
\noindent\textbf{Proxy Task.~}
The evaluation of loss functions requires network training, which costs the most time in the search process.
Similar to other AutoML works~\cite{liu2020evolving,real2020automl}, to accelerate the search process, lightweight proxy tasks for network training are employed in the loss function evaluation. Specifically, fewer training iterations, smaller models and down-sampled images are adopted (see Section~\ref{Sec:Results} and Appendix~\ref{Appendix:Details} for details).
We further improve the efficiency by stopping the network training with invalid loss values (\ie, $\mathrm{NaN}$ and $\mathrm{Inf}$ values).

\section{Experiments}
\label{Sec:Results}

We evaluate \Name\ on different computer vision tasks, including semantic segmentation (Section~\ref{Sec:Seg}), object detection (Section~\ref{Sec:Det}), instance segmentation (Section~\ref{Sec:Msk}), and pose estimation (Section~\ref{Sec:Pse}). The search efficiency of \Name\ is ablated in Section~\ref{Sec:Efficiency}.

\vspace{0.5em}
\noindent\textbf{Implementation Details.~} For the evolutionary algorithm, the population is initialized with $K=20$ randomly generated loss functions, and is restricted to most recent $P=2500$ loss functions. The ratio of tournament selection~\cite{goldberg1991comparative} is set as $T=5\%$ of current population. 
During random initialization and mutations, the sampling probabilities for all the operators in Table~\ref{Methods:Operators} are the same. The initial depth of computational graphs is $D=3$. For the loss-rejection protocol and the gradient-equivalence-check strategy, $B=5$ samples are randomly selected from the training set $\mathcal{S}_{\text train}$. The search and re-training experiments are conducted on 4 NVIDIA V100 GPUs. More details are in Appendix~\ref{Appendix:Details}.

\subsection{Semantic Segmentation}
\label{Sec:Seg}
\paragraph{Settings.} Semantic segmentation concerns categorizing each pixel in an image into a specific class. PASCAL VOC 2012~\cite{everingham2015pascal} with extra annotations~\cite{hariharan2011semantic} is utilized for our experiments. The target evaluation metrics include Mean IoU (mIoU), Frequency Weighted IoU (FWIoU), Global Accuracy (gAcc), Mean Accuracy (mAcc), Boundary IoU (BIoU)~\cite{kohli2009robust} and Boundary F1 Score (BF1)~\cite{csurka2013good}. The first four metrics measure the overall segmentation accuracy, and the other two metrics evaluate the boundary accuracy. 

During search, we use DeepLabv3+~\cite{chen2018encoder} with ResNet-50~\cite{he2016deep} as the network. Following \cite{li2020auto}, we simplify the proxy task by down-sampling the input images to the resolution of 128$\times$128, and reducing the training schedule to 3 epochs ($1/10$ of the normal training schedule).
After the search procedure, we re-train the segmentation networks with ResNet-101~\cite{he2016deep} as the backbone for 30 epochs. The input image resolution is 512$\times$512. The re-training setting is the same as \cite{chen2018encoder}, except that the searched loss function is utilized. More details are in Appendix~\ref{Appendix:Seg}.

\vspace{0.5em}
\noindent\textbf{Results.~} Table~\ref{Exp:Seg} compares our searched losses with the widely used cross-entropy loss, other metric-specific handcrafted loss functions, and the surrogate losses searched by ASL~\cite{li2020auto}. Note that ASL is restricted to a specific designed search space for semantic segmentation, which cannot be simply extended to handle generic metrics. The results show that our searched losses outperform the manually designed losses consistently, and on par with or better than the searched losses by ASL on most target metrics. Only for the BIoU metric, our searched loss is slightly worse than that of ASL, but is still much better than the handcrafted losses. 
Appendix~\ref{Appendix:Losses} presents the formulas of the discovered loss functions, which indicate that the intermediate aggregations (\eg, Max-Pooling$_{3\times3}$ and Mean$_{nhw}$) between non-linear operations may have potential benefits for metrics such as mAcc, BIoU, and BF1.

\vspace{0.5em}
\noindent\textbf{Generalization of the searched functions.~} To verify the generalization ability of the searched losses, we conduct re-training experiments on different datasets and networks using the CE loss and the losses originally searched for DeepLabv3+~\cite{chen2018encoder} with ResNet50~\cite{he2016deep} on PASCAL VOC~\cite{everingham2015pascal}. Due to limited computational resource, we only compare on mIoU and BF1 metrics. Table~\ref{Exp:Generalization.Seg} summarizes the results on PASCAL VOC and Cityscapes~\cite{cordts2016cityscapes}, using DeepLabv3+~/~PSPNet~\cite{zhao2017pyramid} with ResNet-50~/~ResNet-101 as the networks. The results show that the searched loss functions generalize well between different datasets, and can be applied to various semantic segmentation networks.

\setlength{\tabcolsep}{1.5pt}
\renewcommand{\arraystretch}{1.2}
\begin{table}[t]
\begin{center}
\small
\resizebox{0.83\linewidth}{!}{
\begin{tabular}{l|l|cccccc}
\hline
\multicolumn{2}{c|}{\bf Loss Function} & \multicolumn{1}{c}{\bf \small mIoU} & \multicolumn{1}{c}{\bf \small FWIoU} & \multicolumn{1}{c}{\bf \small gAcc} & \multicolumn{1}{c}{\bf \small mAcc} & \multicolumn{1}{c}{\bf \small BIoU} & \multicolumn{1}{c}{\bf \small BF1}\TBstrut \\
\hline
\multicolumn{2}{l|}{Cross Entropy} & 78.7 & 91.3 & \underline{\bf 95.2} & 87.3 & 70.6 & 65.3\Tstrut \\
\multicolumn{2}{l|}{WCE~\cite{ronneberger2015u}} & 69.6 & 85.6 & 91.1 & \underline{\bf 92.6} & 61.8 & 37.6\\
\multicolumn{2}{l|}{DiceLoss~\cite{milletari2016v}} & 77.8 & 91.3 & 95.1 & 87.5 & 69.9 & 64.4\\
\multicolumn{2}{l|}{Lov{\`a}sz~\cite{berman2018lovasz}} & \underline{79.7} & \bf 91.8 & \bf 95.4 & 88.6 & 72.5 & 66.7\\
\multicolumn{2}{l|}{DPCE~\cite{caliva2019distance}} & 79.8 & \bf 91.8 & \bf 95.5 & 87.8 & \underline{71.9} & \underline{66.5}\\
\multicolumn{2}{l|}{SSIM~\cite{qin2019basnet}} & 79.3 & \bf 91.7 & \bf 95.4 & 87.9 & \underline{71.5} & \underline{66.4}\Bstrut\\
\hline
\multirow{2}{*}{mIoU} & ASL~\cite{li2020auto}~ & \underline{\bf 81.0} & \bf{92.1} & \bf 95.7 & 88.2 & 73.4 & 68.9\Tstrut\\
                      & Ours & \underline{\bf 80.7} & \bf{92.1} & \bf{95.7} & 89.1 & 74.1 & 66.0\Bstrut\\
 \hline
 \multirow{2}{*}{FWIoU} & ASL~\cite{li2020auto} & 80.0 & \underline{\bf 91.9} & \bf 95.4 & 89.2 & 75.1 & 65.7\Tstrut\\
                        & Ours & 78.7 & \underline{\bf 91.7} & {\bf 95.2} & 87.7 & 72.9 & 64.6\Bstrut\\
 \hline
 \multirow{2}{*}{gAcc} & ASL~\cite{li2020auto} & 79.7 & \bf 91.8 & \underline{\bf 95.5} & 89.0 & 74.1 & 64.4\Tstrut\\
                       & Ours & 79.4 & {\bf 91.7} & \underline{\bf 95.3} & 88.7 & 73.6 & 64.8\Bstrut\\
 \hline
 \multirow{2}{*}{mAcc} & ASL~\cite{li2020auto} & 69.8 & 85.9 & 91.3 & \underline{\bf 92.7} & 72.9 & 35.6\Tstrut\\
                      & Ours & 75.3 & 89.2 & 93.7 & \underline{\bf 92.6} & 73.7 & 44.1\Bstrut\\
 \hline
 \multirow{2}{*}{BIoU} & ASL~\cite{li2020auto} & 49.0 & 69.9 & 62.6 & 81.3 & \underline{\bf 79.2} & 39.0\Tstrut\\
                      & Ours & 39.8 & 66.6 & 79.7 & 47.8 & \underline{77.6} & 45.5\Bstrut\\
 \hline
 \multirow{2}{*}{BF1} & ASL~\cite{li2020auto} & 1.9 & 1.0 & 2.7 & 6.5 & 7.4 & \underline{74.8}\Tstrut\\
                      & Ours & 6.0 & 54.6 & 73.8 & 7.3 & 9.4 & \underline{\bf 79.8}\Bstrut\\
 \hline
\end{tabular}}
\end{center}
\caption{Semantic segmentation results of DeepLabv3+~\cite{chen2018encoder} with ResNet-101~\cite{he2016deep} on PASCAL VOC~\cite{everingham2015pascal}. Results of the target metric(s) for each loss function are \underline{underlined}, and the highest results within a tolerance of 0.5 are in \textbf{bold}.}
\label{Exp:Seg}
\end{table}

\setlength{\tabcolsep}{3pt}
\renewcommand{\arraystretch}{1.2}
\begin{table}[t]
\begin{center}
\small
\resizebox{0.85\linewidth}{!}{
\begin{tabular}{l|l|cc|cc|cc}
\hline
\multicolumn{2}{c|}{\bf Dataset} & \multicolumn{2}{c|}{\bf Cityscapes} & \multicolumn{4}{c}{\bf VOC} \\
\hline
\multicolumn{2}{c|}{\bf Network} & \multicolumn{2}{c|}{\bf R101-DLv3+} & \multicolumn{2}{c|}{\bf R50-DLv3+} & \multicolumn{2}{c}{\bf R101-PSP}\TBstrut\\
\hline
\multicolumn{2}{c|}{\bf Loss Function} & \multicolumn{1}{c}{\bf mIoU} & \multicolumn{1}{c|}{\bf BF1} & \multicolumn{1}{c}{\bf mIoU} & \multicolumn{1}{c|}{\bf BF1} & \multicolumn{1}{c}{\bf mIoU} & \multicolumn{1}{c}{\bf BF1}\TBstrut\\
\hline
\multicolumn{2}{c|}{Cross Entropy} & 80.0 & 62.2 & 76.2 & 61.8 & 77.9 & 64.7\TBstrut\\
\hline
\multirow{2}{*}{mIoU} & ASL~\cite{li2020auto}~ & \underline{\bf 80.7} & 66.5 & \underline{\bf 78.4} & 66.9 & \underline{\bf 78.9} & 65.7\Tstrut\\
 & Ours & \underline{\bf 80.4} & 63.8 & \underline{\bf 78.0} & 62.8 & \underline{\bf 78.5} & 64.9\Bstrut\\
\hline
\multirow{2}{*}{BF1} & ASL~\cite{li2020auto}~ & 6.7 & \underline{\bf 78.0} & 1.4 & \underline{70.8} & 1.6 & \underline{\bf 71.8}\Tstrut\\
 & Ours & 9.3 & \underline{\bf 77.7} & 7.2 & \underline{\bf 78.3} & 5.1 & \underline{\bf 71.3}\Bstrut\\
\hline
\end{tabular}}
\end{center}
\caption{Generalization of the searched loss functions for semantic segmentation among different datasets and networks. The losses are originally searched for DeepLabv3+~\cite{chen2018encoder} with ResNet-50~\cite{he2016deep} on PASCAL VOC~\cite{everingham2015pascal}. ``R50'' and ``R101'' are the abbreviations of ResNet-50 and ResNet-101, respectively. ``DLv3+'' and ``PSP'' denote the DeepLabv3+ and PSPNet~\cite{zhao2017pyramid}, respectively.}
\label{Exp:Generalization.Seg}
\vspace{-1em}
\end{table}

\subsection{Object Detection}
\label{Sec:Det}
\paragraph{Settings.} Object detection is the task of detecting the bounding boxes and categories of instances belonging to certain classes. To evaluate our algorithm, we conduct experiments on the widely used COCO dataset~\cite{lin2014microsoft}. The target evaluation metric is Mean Average Precision (mAP). 

We use Faster R-CNN~\cite{ren2015faster} with ResNet-50~\cite{he2016deep} and FPN~\cite{lin2017feature} as the detection network. There are 4 loss branches, \ie, the classification and regression branches for the RPN~\cite{ren2015faster} sub-network and Fast R-CNN~\cite{girshick2015fast} sub-network. We search for loss functions of the 4 branches simultaneously from scratch. Following \cite{rezatofighi2019generalized}, we use the intersection, union and enclosing areas between the predicted and ground-truth boxes as the regression loss inputs.

During the search, we train the network with $1/4$ of the COCO data for 1 epoch as the proxy task. We further simplify the network by only using the last three feature levels of FPN, and reducing the channels of the detection head by half. After the search procedure, we re-train the detection network with the searched loss functions. The re-training hyper-parameters are the same as the default settings of MMDetection~\cite{chen2019mmdetection}. More details are in Appendix~\ref{Appendix:Det}.

\vspace{0.5em}
\noindent\textbf{Results.~} Table~\ref{Exp:Det} compares our searched loss functions with the handcrafted loss functions and the searched function by \cite{liu2021loss}. The effectiveness of our method is verified for searching on only the 2 branches of the Fast R-CNN~\cite{girshick2015fast} sub-network, and on all of the 4 branches. Results show that our searched losses are on par with the existing handcrafted and searched loss functions. Note that \cite{liu2021loss} designs the search space and strategies specifically for object detection, and initializes their search process with handcrafted loss functions, while \Name\ is a general framework that searches for loss functions from scratch.
The formulas of the discovered loss functions are presented in Appendix~\ref{Appendix:Losses}. The searched loss function for bounding box regression shares a similar expression with the GIoULoss~\cite{rezatofighi2019generalized}, confirming the effectiveness of the handcrafted loss function.

\vspace{0.5em}
\noindent\textbf{Generalization of the searched functions.~} We verify the generalization ability of the searched loss function in Table~\ref{Exp:Generalization.Det}. The loss is originally searched on COCO~\cite{lin2014microsoft} with ResNet-50~\cite{he2016deep}, and is used for training networks with different backbone (\ie, ResNet-101) and on different dataset (\ie, PASCAL VOC~\cite{everingham2015pascal}). The results show that our searched loss functions can generalize well to different object detection networks and datasets.

\setlength{\tabcolsep}{3pt}
\renewcommand{\arraystretch}{1.2}
\begin{table}[t]
\begin{center}
\small
\resizebox{0.73\linewidth}{!}{
\begin{tabular}{cc|cc|c}
\hline
\multicolumn{4}{c|}{\bf Loss Function} & \multirow{2}{*}{\bf mAP}\TBstrut \\
\cline{1-4}
\bf Cls$_\text{RPN}$ & \bf Reg$_\text{RPN}$~ & \bf ~Cls$_\text{RCNN}$ & \bf Reg$_\text{RCNN}$ & \TBstrut \\
\hline
CE & L1 & CE & L1 & 37.3\Tstrut \\
CE & L1 & CE & IoULoss~\cite{yu2016unitbox} & 37.9 \\
CE & L1 & CE & GIoULoss~\cite{rezatofighi2019generalized} & 37.6 \\
CE & L1 & \multicolumn{2}{c|}{CSE-Auto-A~\cite{liu2021loss}} & 38.5\Bstrut \\
\hline
CE & L1 & \multicolumn{2}{c|}{Ours} & 38.0\TBstrut \\
\cline{1-4}
\multicolumn{4}{c|}{Ours} & 38.1\TBstrut \\
\hline
\end{tabular}}
\end{center}
\caption{Object detection results of ResNet-50~\cite{he2016deep} on COCO~\cite{lin2014microsoft}. $\mathrm{Cls}$ and $\mathrm{Reg}$ are the classification and regression branches, respectively, where the subscripts $\mathrm{RPN}$ and $\mathrm{RCNN}$ denote the RPN~\cite{ren2015faster} sub-network and Fast R-CNN~\cite{girshick2015fast} sub-network, respectively.}
\label{Exp:Det}
\end{table}

\setlength{\tabcolsep}{3pt}
\renewcommand{\arraystretch}{1.2}
\begin{table}[t]
\begin{center}
\small
\resizebox{0.8\linewidth}{!}{
\begin{tabular}{c|c|c}
\hline
\multicolumn{1}{c|}{\bf Dataset} & \multicolumn{1}{c|}{\bf COCO} & \multicolumn{1}{c}{\bf VOC}\TBstrut\\
\hline
\multicolumn{1}{c|}{\bf Network} & \multicolumn{1}{c|}{\bf ResNet-101} & \multicolumn{1}{c}{\bf ResNet-50}\TBstrut\\
\hline
\multicolumn{1}{c|}{\bf Loss Function} & \multicolumn{1}{c|}{\bf mAP} & \multicolumn{1}{c}{\bf mAP}\TBstrut\\
\hline
CE + L1 + CE + IoULoss~\cite{yu2016unitbox} & 39.7 & 80.4\Tstrut\\
Ours & 39.9 & 80.6\Bstrut\\
\hline
\end{tabular}
}
\end{center}
\caption{Generalization of the searched losses for object detection among different datasets and networks. The loss is originally searched for ResNet-50~\cite{he2016deep} on COCO~\cite{lin2014microsoft}.}
\label{Exp:Generalization.Det}
\end{table}

\subsection{Instance Segmentation}
\label{Sec:Msk}
\paragraph{Settings.} Instance segmentation is the task of detecting the segmentation masks and categories of instances. We also conduct experiments on COCO~\cite{lin2014microsoft}, except that the target metric is mAP with IoU defined on masks.

Mask R-CNN~\cite{he2017mask} with ResNet-50~\cite{he2016deep} and FPN~\cite{lin2017feature} is used as the network. We conduct the search for all the 5 loss branches simultaneously. The proxy task is the same as for object detection. We use the default hyper-parameters of MMDetection~\cite{chen2019mmdetection} for re-training the networks with our searched loss functions. More details are in Appendix~\ref{Appendix:Ins}.

\vspace{0.5em}
\noindent\textbf{Results.~} Table~\ref{Exp:Msk_Pose}~(a) summarizes the results. The loss function searched from scratch by \Name\ is on par with the existing manually designed loss functions. The discovered loss functions are presented in Appendix~\ref{Appendix:Losses}.

\setlength{\tabcolsep}{3pt}
\renewcommand{\arraystretch}{1.2}
\begin{table}[t]
\begin{subtable}[b]{0.58\linewidth}
\centering
\small
\resizebox{1\linewidth}{!}{
    \begin{tabular}{l|c}
    \hline
    \multicolumn{1}{c|}{\bf Loss Function} & \multirow{1}{*}{\bf ~mAP}\TBstrut \\
    \hline
    \footnotesize CE + L1 + CE + L1 + CE & 34.6\Tstrut \\
    \footnotesize CE + L1 + CE + IoULoss~\cite{yu2016unitbox} + CE & 34.4 \\
    \footnotesize CE + L1 + CE + GIoULoss~\cite{rezatofighi2019generalized} + CE & 34.7\Bstrut \\
    \hline
    \multicolumn{1}{c|}{Ours} & 34.8\TBstrut \\
    \hline
    \end{tabular}}
    \caption{Instance Segmentation}
    \label{Exp:Msk}
\end{subtable}
\hspace{\fill}
\begin{subtable}[b]{0.36\linewidth}
\centering
\small
\resizebox{0.9\linewidth}{!}{
    \begin{tabular}{c|c}
    \hline
    \multicolumn{1}{c|}{\bf Loss Function} & \multicolumn{1}{c}{\bf mAP}\TBstrut \\
    \hline
    MSE & 71.5\TBstrut\\
    \hline
    Ours & 72.0\TBstrut\\
    \hline
    \end{tabular}}
    \vspace{1.5em}
    \caption{Pose Estimation}
    \label{Exp:Pose}
\end{subtable}
\vspace{0.1em}
\caption{Instance segmentation and pose estimation results of ResNet-50~\cite{he2016deep} on COCO~\cite{lin2014microsoft}. In the first three rows of (a), the five losses correspond to the Cls$_\text{RPN}$, Reg$_\text{RPN}$, Cls$_\text{RCNN}$, Reg$_\text{RCNN}$ and Mask branches, respectively. ``MSE'' in (b) denotes the mean square error loss used by \cite{xiao2018simple}.}
\label{Exp:Msk_Pose}
\vspace{-1.0em}
\end{table}

\subsection{Pose Estimation}
\label{Sec:Pse}
\paragraph{Settings.} Pose estimation is the task of localizing the keypoints of humans. Experiments are conducted on COCO~\cite{lin2014microsoft}. The target evaluation metric is keypoint mAP, which is very similar to the mAP in object detection, where object keypoint similarity (OKS) is used to substitute the role of bounding box IoU.

We use \cite{xiao2018simple} with Resnet-50~\cite{he2016deep} as the network. Following the practice in \cite{mmpose2020}, person detection results provided by \cite{xiao2018simple} are utilized. 
During search, we train the network for 4 epochs as the proxy task. After the search procedure, we re-train the network with the searched loss functions using the default training settings of MMPose~\cite{mmpose2020}.  More details are in Appendix~\ref{Appendix:Pose}.

\vspace{0.5em}
\noindent\textbf{Results.~} Table~\ref{Exp:Msk_Pose}~(b) compares the results of our searched loss function with the widely used mean square error (MSE) loss. Starting from randomly initialized loss functions, our searched loss function is slightly better than the MSE loss, demonstrating the effectiveness of \Name. 
Appendix~\ref{Appendix:Losses} presents the formulas of the discovered loss functions. The searched function learns a regularization term to punish too large prediction values.

\begin{figure}[t]
    \centering
    \begin{subfigure}{0.48\linewidth}
        \includegraphics[width=\textwidth]{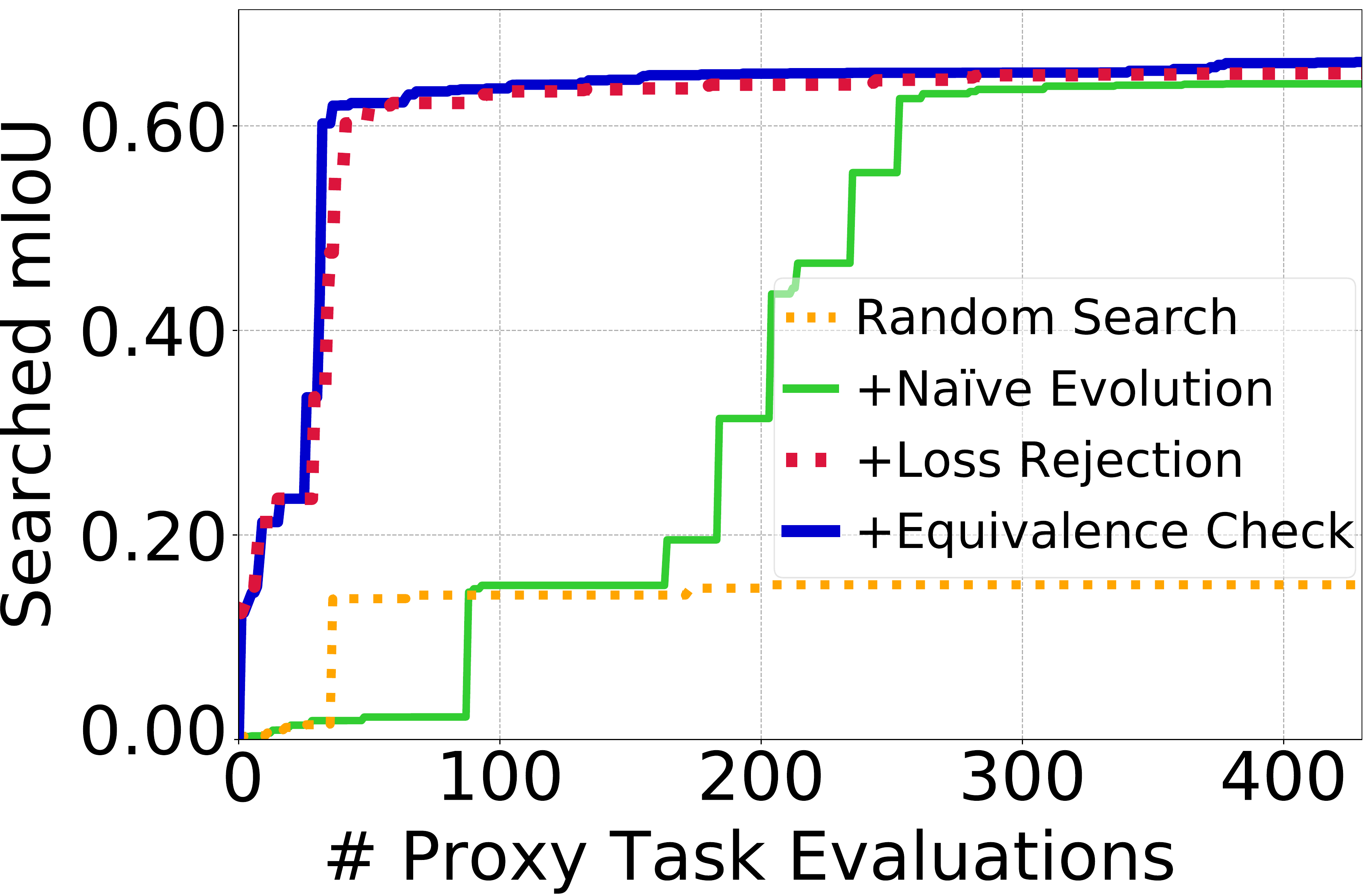}
        \vspace{-1.4em}
        \caption{Semantic Segmentation}
    \end{subfigure}
    \hspace{\fill}
    \begin{subfigure}{0.48\linewidth}
        \includegraphics[width=\textwidth]{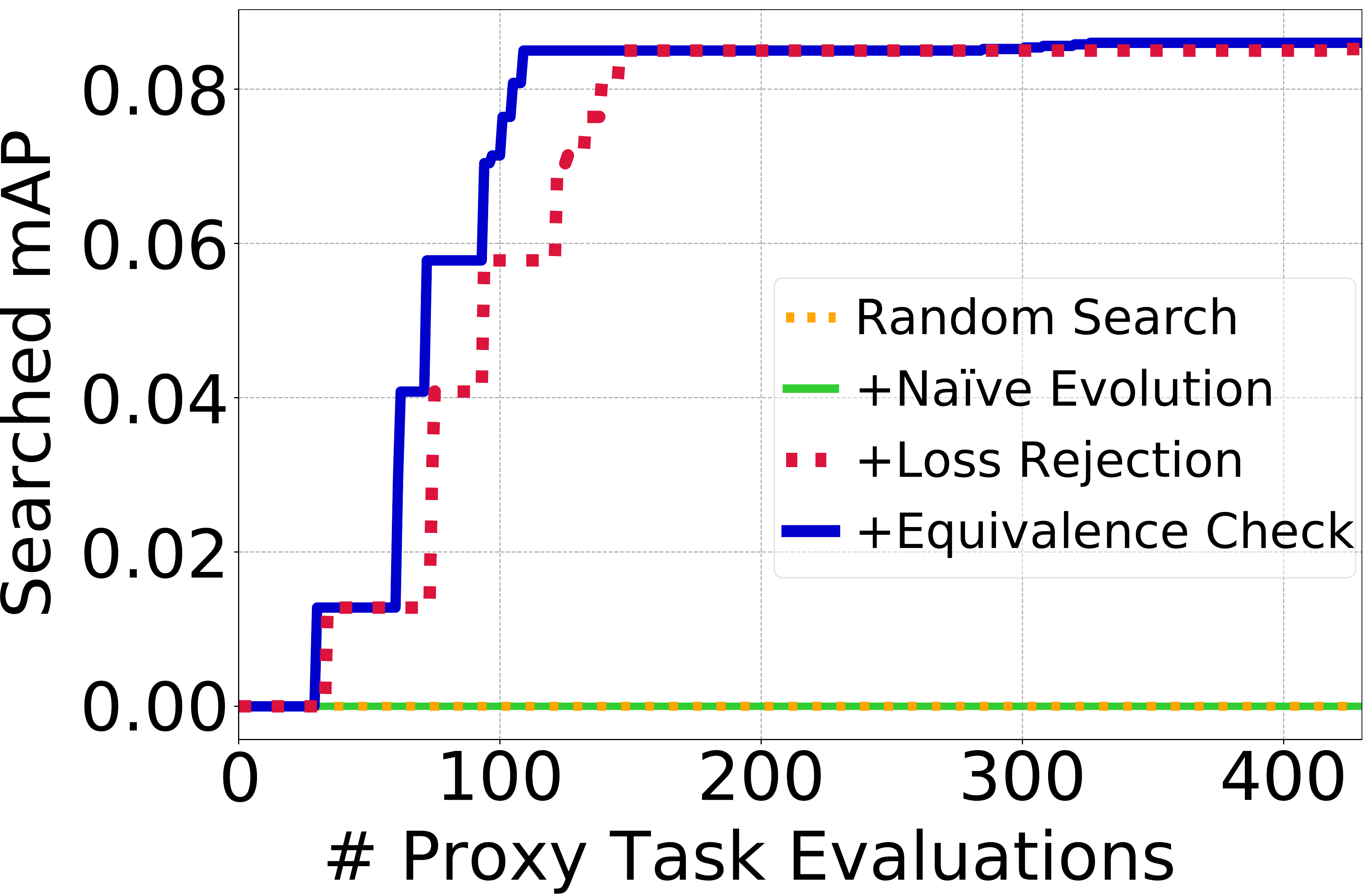}
        \vspace{-1.4em}
        \caption{Object Detection}
    \end{subfigure}
    \vspace{0.3em}
    \caption{Ablation study on search efficiency of \Name. Each curve presents the averaged evaluation scores over the top-5 loss functions in the current population.}
    \vspace{-1.2em}
    \label{Fig:Exp.Efficiency}
\end{figure}

\setlength{\tabcolsep}{3pt}
\renewcommand{\arraystretch}{1.2}
\begin{table}[b]
\begin{center}
\small
\resizebox{0.95\linewidth}{!}{
\begin{tabular}{l|c|c}
\hline
 & {\bf Speed-Up} & {\bf \# Explored Losses}\TBstrut\\
\hline
Na\"ive Evolution & 1$\times$ & $\sim$300\Tstrut\\
+ Loss-Rejection Protocol & $\sim$700$\times$ & $\sim$2.1$\times$10$^5$ \\
+ Gradient-Equivalence-Check Strategy & $\sim$1000$\times$ & $\sim$3.2$\times$10$^5$ \\
+ $^\dag$Stop Training for Invalid Loss Values & $\sim$5000$\times$ & $\sim$1.5$\times$10$^6$\Bstrut\\
\hline
\end{tabular}}
\end{center}
\caption{Search speed of degenerated variants of \Name\ on object detection. ``\# Explored Losses'' demonstrates the number of losses that can be explored in 48 hours. $\dag$ ``Stop Training for Invalid Loss Values'' means that the network training is stopped in the first 20 iterations due to invalid loss values (\ie, $\mathrm{NaN}$ and $\mathrm{Inf}$).}
\label{Tab:Exp.Efficiency}
\end{table}

\subsection{Search Efficiency}
\label{Sec:Efficiency}

We ablate the search efficiency of \Name\ on semantic segmentation with the mIoU metric and object detection with the mAP metric. Starting from random search, we add each component of \Name\ sequentially to verify its effectiveness. Figure~\ref{Fig:Exp.Efficiency} shows the results. The proposed loss-rejection protocol greatly improves the search efficiency for both tasks, and the gradient-equivalence-check strategy is also helpful. For semantic segmentation which only contains a single loss branch, na\"ive evolution can also discover the proper losses despite its inefficiency. For object detection with 4 loss branches, the exponentially increased sparsity of the search space brings great difficulty to the search. Therefore, without the loss-rejection protocol, no loss functions with scores greater than zero can be discovered within a reasonable time. Table~\ref{Tab:Exp.Efficiency} further presents the number of loss functions that can be explored in 48 hours by \Name. Over $10^6$ loss functions can be explored, which ensures that \Name\ can explore the huge and sparse search space within a reasonable time.

\section{Conclusion}
\label{Sec:Conclusion}

\Name\ is the first general framework for searching loss functions from scratch for generic tasks. The search space is composed only of basic primitive operators. A variant of evolutionary algorithm is employed for searching, where a loss-rejection protocol and a gradient-equivalence-check strategy are developed to improve the search efficiency.
\Name\ can discover loss functions that are on par with or superior to existing loss functions on various tasks with minimal human expertise.

\paragraph{Acknowledgments} The work is supported by the National Key R\&D Program of China (2020AAA0105200), Beijing Academy of Artificial Intelligence and the Institute for Guo Qiang of Tsinghua University.

{\small
\bibliographystyle{ieee_fullname}
\bibliography{egbib}
}

\clearpage
\appendix
\section{Implementation Details}
\label{Appendix:Details}

For the evolutionary algorithm, the population is initialized with $K=20$ randomly generated loss functions, and is restricted to most recent $P=2500$ loss functions. The ratio of tournament selection~\cite{goldberg1991comparative} is set as $T=5\%$ of current population. 
During random initialization and mutations, the sampling probabilities for all the operators in the primitive operator set $\mathcal{H}$ are the same. The initial depth of computational graphs is $D=3$. For the loss-rejection protocol and the gradient-equivalence-check strategy, $B=5$ samples are randomly selected from the training set $\mathcal{S}_{\text train}$.

In the proposed loss-rejection protocol, the threshold for unpromising loss function is set as $\eta=0.6$. We use stochastic gradient descent with momentum to optimize Eq.~\eqref{Eq:Method.EarlyReject} for $500$ iterations. The learning rate for the loss-rejection protocol is $0.001$. The momentum factor is set as $0.9$. There is no weight decay. Since the predictions of different samples and spatial positions are optimized independently in Eq.~\eqref{Eq:Method.EarlyReject}, we do not normalize along any dimensions when aggregating the output tensor to the final loss value in the loss-rejection protocol.

We stop the search when the total number of proxy task evaluations reaches $500$. The best loss function is then used for re-training experiments. All the experiments are conducted on 4 NVIDIA V100 GPUs.

\begin{table*}[ht]
\begin{center}
\small
\resizebox{0.85\linewidth}{!}{
\begin{tabular}{l|l|lc}
\hline
\bf Task & \bf Metric & \multicolumn{2}{c}{\bf Formula\TBstrut}\\
\hline
\multirow{6}{*}{\vspace{-0.7cm}Seg} & mIoU & \multicolumn{2}{c}{$-\frac12\log\left(\tanh\left(\frac{y}{1+y}\sqrt{\hat y}\right)(\hat{y}^2+y)\right) $}\Tstrut\\
& FWIoU & \multicolumn{2}{c}{$\log(y\log\left(y + (y + \hat y) / \hat y\right))$}\\
& gAcc & \multicolumn{2}{c}{$\exp\left(\left(y-\sqrt{\hat y}\right)^2\right)$}\\
& mAcc & \multicolumn{2}{c}{$\tanh\left(\sqrt{\mathrm{Mean}_{nhw}\left(-\log(\hat y)\sqrt{\mathrm{Max\text{-}Pooling}\left(\sqrt{2y}\right)}\right)}\right)$}\\
& BIoU & \multicolumn{2}{c}{$-\log\left({\hat y}^3y\sqrt{-\mathrm{Min\text{-}Pooling}\left(-\hat y\right)}\exp\left(\left(\hat y\exp(\hat y)\right)^2\right)\right)$}\\
& BF1 & \multicolumn{2}{c}{$\exp\left(\mathrm{Mean}_{nhw}\left(-\mathrm{Max\text{-}Pooling}(y)\log\left(\mathrm{Max\text{-}Pooling}(\hat y)\right)\right) + \tanh\left(\mathrm{Mean}_{nhw}\left(2\hat y + 1\right)\right)\right)$}\Bstrut\\
\hline
\multirow{4}{*}{Det} & \multirow{4}{*}{mAP} & Cls$_\text{RPN}$ & $\exp(\tanh(y)) / \hat y$\Tstrut\\
& & Reg$_\text{RPN}$ & $e / (i + u)$\\
& & Cls$_\text{RCNN}$ & $-y\log(\hat y)$\\
& & Reg$_\text{RCNN}$ & $- \log(i / e) $\Bstrut\\
\hline
\multirow{5}{*}{Ins} & \multirow{5}{*}{mAP} & Cls$_\text{RPN}$ & $-y\hat y / \tanh(\hat y)$\Tstrut\\
& & Reg$_\text{RPN}$ & $e / (i + u)$\\
& & Cls$_\text{RCNN}$ & $-y\log(\hat y)$\\
& & Reg$_\text{RCNN}$ & $- \sqrt{i} / \sqrt{e} $\\
& & Mask & $|\log(y\hat y)|$\Bstrut\\
\hline
Pose & mAP & \multicolumn{2}{c}{$(\hat{y} - 16y^4)^2+\hat{y}^4$\TBstrut}\\
\hline
\end{tabular}
}
\end{center}
\caption{Discovered Loss Functions. ``Seg'', ``Det'', ``Ins'', and ``Pose'' denotes the tasks of semantic segmentation, object detection, instance segmentation, and pose estimation, respectively. $\mathrm{Max\text{-}Pooling}$ and $\mathrm{Min\text{-}Pooling}$ have kernel size $3\times3$. A small positive number $\epsilon=10^{-12}$ is added in $\log(\cdot), \sqrt{\cdot}$, and the denominators of division operations to avoid infinite values or gradients. $\hat{y}$ and $y$ denote the network prediction and the training target of the corresponding branch. $i$, $u$, and $e$ in the box regression loss branches refer to the intersection, union, and enclosed areas between the predicted and the target bounding boxes, respectively, which follows \cite{liu2021loss}.}\vspace{-0.5em}
\label{Appendix:Formulae}
\end{table*}

\subsection{Semantic Segmentation}
\label{Appendix:Seg}

\noindent\textbf{Datasets.~} PASCAL VOC 2012~\cite{everingham2015pascal} with extra annotations~\cite{hariharan2011semantic} is utilized for our experiments. During search, we randomly sample $1500$ training images in
PASCAL VOC to form the evaluation set $\mathcal{S}_{\text{eval}}$, and use the remaining training images as the training set $\mathcal{S}_{\text{train}}$. The target evaluation metrics include Mean IoU (mIoU), Frequency Weighted IoU (FWIoU), Global Accuracy (gAcc), Mean Accuracy (mAcc), Boundary IoU (BIoU)~\cite{kohli2009robust} and Boundary F1 Score (BF1)~\cite{csurka2013good}. The first four metrics measure the overall segmentation accuracy, and the other two metrics evaluate the boundary accuracy.

\vspace{0.5em}
\noindent\textbf{Implementation Details.~}
During search, we use DeepLabv3+~\cite{chen2018encoder} with ResNet-50~\cite{he2016deep} as the network. The softmax probabilities $y$ and the one-hot ground-truth labels $\hat{y}$ are used as the inputs of loss functions, both of which are of shape $(N,C,H,W)$. For the boundary metrics, we use the pre-computed boundaries of $y$ as the training targets. Following \cite{li2020auto}, we simplify the proxy task by down-sampling the input images to the resolution of $128\times128$, and reducing the training schedule to $3$ epochs ($1/10$ of the normal training schedule) with a mini-batch size of $32$. We use stochastic gradient decent with momentum to train the network. The initial learning rate is $0.02$, which is decayed by polynomial with power $0.9$ and minimum learning rate $10^{-4}$. The momentum and weight decay factors are set to $0.9$ and $5\times10^{-4}$, respectively. For faster convergence, the learning rate of the segmentation head is multiplied by $10$. After the search procedure, we re-train
the segmentation networks with ResNet-101 as the backbone for $30$ epochs.  The input image resolution is $512\times512$. The re-training setting is the same as \cite{chen2018encoder}, except that the searched loss function is utilized.

\subsection{Object Detection}
\label{Appendix:Det}

\noindent\textbf{Datasets.~} We conduct experiments on large-scale object detection dataset COCO~\cite{lin2014microsoft}. In the search experiments, we randomly sample $5000$ images from the
training set for validation purpose, and sample $1/4$ of the remaining images for network training. The target evaluation metric is Mean Average Precision (mAP). 

\vspace{0.5em}
\noindent\textbf{Implementation Details.~} We use Faster R-CNN~\cite{ren2015faster} with ResNet-50~\cite{he2016deep} and FPN~\cite{lin2017feature} as the detection network. There are 4 loss branches, \ie, the classification and regression branches for the RPN~\cite{ren2015faster} sub-network and Fast R-CNN~\cite{girshick2015fast} sub-network. We search for loss functions of the 4 branches simultaneously from scratch. The inputs of loss functions for the classification branches are the softmax probabilities and the one-hot ground-truth labels. Following \cite{liu2021loss}, we use the intersection, union and enclosing areas between the predicted and ground-truth boxes as the regression loss inputs. The loss weights are set to $1.0$ for classification branches and $10.0$ for regression branches, which follows \cite{rezatofighi2019generalized}.

During search, the initialization / mutation process is repeated for each loss branch separately until it passes the loss-rejection protocol. In the loss-rejection process of each loss branch, the predictions of the other branches are set as the ground-truth targets. For the RPN sub-network, the correlation score $g(L; \xi)$ is calculated with the predicted region proposals and the corresponding training targets.

We train the network with $1/4$ of the COCO data for $1$ epoch as the proxy task. We further simplify the network by only using the last three feature levels of FPN, and reducing the channels of the detection head by half. The learning rate is set to $0.04$ with a batch size of $32$, and a linear learning rate warm-up of $250$ iterations is applied. After the search procedure, we re-train the detection network with the searched loss functions for $12$ epochs. The re-training hyper-parameters are the same as the default settings of MMDetection~\cite{chen2019mmdetection}.

\subsection{Instance Segmentation}
\label{Appendix:Ins}

\noindent\textbf{Datasets.~} We conduct experiments on COCO~\cite{lin2014microsoft}. In the search experiments, we randomly sample $5000$ images from the
training set for validation purpose, and sample $1/4$ of the remaining images for network training. The target metric is mAP with IoU defined on masks.

\vspace{0.5em}
\noindent\textbf{Implementation Details.~} 
Mask R-CNN~\cite{he2017mask} with ResNet-50~\cite{he2016deep} and FPN~\cite{lin2017feature} is used as the network. We conduct the search for all the 5 loss branches simultaneously. The inputs of loss functions for the classification and regression branches are the same as in object detection. The prediction used as loss inputs for each RoI of the mask branch is the per-pixel softmax probabilities indicating foreground or background.

The proxy task is the same as for object detection. After the search procedure, we re-train the detection network with the searched loss functions for $12$ epochs. We use the default hyper-parameters of MMDetection~\cite{chen2019mmdetection} for re-training the networks with our searched loss functions.

\subsection{Pose Estimation}
\label{Appendix:Pose}

\noindent\textbf{Datasets.~}
Experiments are conducted on COCO~\cite{lin2014microsoft}, which contains 250,000 person instances labeled with 17 keypoints each. During search, we randomly select $5000$ image from the training set for proxy task evaluation. The target evaluation metric is keypoint mAP, which is very similar to the mAP in object detection, where object keypoint similarity (OKS) is used to substitute the role of bounding box IoU.

\vspace{0.5em}
\noindent\textbf{Implementation Details.~} 
We use \cite{xiao2018simple} with Resnet-50~\cite{he2016deep} as the network. Following the practice in \cite{mmpose2020}, person detection results provided by \cite{xiao2018simple} are utilized. For each joint of each detected box in the person detection results, the network predicts a $64\times48$ heatmap, and the target heatmap is constructed as in \cite{xiao2018simple}. Loss function is applied only on visible joints.

During search, we train the network using the Adam~\cite{kingma2014adam} optimizer for $4$ epochs as the proxy task. The learning rate is $2.5\times 10^{-4}$ with a cosine annealing schedule and a linear warm-up for $250$ iterations. The batch size is set as $256$. After the search procedure, we re-train the network with the searched loss functions for $210$ epochs using the default training settings of MMPose~\cite{mmpose2020}.

\section{Formulas of the Discovered Loss Functions}
\label{Appendix:Losses}
Table \ref{Appendix:Formulae} shows the formulas of the discovered loss functions. The formulas are simplified manually by removing useless operators (\eg squaring the constant one or taking the absolute value of a non-negative quantity).

\end{document}